\pdfoutput=1
\documentclass[11pt]{article}
\usepackage[final]{acl} %

\usepackage{times}
\usepackage{latexsym}
\usepackage[T1]{fontenc}
\usepackage[utf8]{inputenc}
\usepackage{microtype}
\usepackage{inconsolata}
 
\usepackage{amsmath}
\usepackage{amsthm}
\usepackage{amssymb}
\usepackage{amsthm}
\usepackage{subfigure}

\usepackage{graphicx} %
\usepackage{xcolor,colortbl}
\usepackage{url}
\usepackage{physics}
\usepackage{tikz-cd}
\usepackage{caption}
\usepackage{hyperref}

\renewcommand{\eqref}[1]{Equation~\ref{eq:#1}}
\newcommand{\secref}[1]{Section~\ref{sec:#1}}
\newcommand{\secstworef}[2]{Sections~\ref{sec:#1} and~\ref{sec:#2}}

\newcommand{\appref}[1]{Appendix~\ref{app:#1}}

\newcommand{\figref}[1]{Figure~\ref{fig:#1}}
\newcommand{\figstworef}[2]{Figures~\ref{fig:#1} and~\ref{fig:#2}}

\newcommand{\promptref}[1]{Prompt~\ref{pro:#1}}
\newcommand{\promptstworef}[2]{Prompts~\ref{pro:#1} and~\ref{pro:#2}}

\newcommand{\tabref}[1]{Table~\ref{tab:#1}}

\newcommand{\arxiv}[0]{arXiv}

\usepackage[framemethod=TikZ]{mdframed}
\usepackage{xcolor}
\usepackage{listings}
\usepackage{caption}
\newmdenv[%
    innertopmargin=-3pt,
    innerbottommargin=-3pt,
    innerleftmargin=3pt,
    innerrightmargin=-2pt,
    skipabove=0pt,
    skipbelow=0pt,
    backgroundcolor=gray!10,
    linecolor=black,
    outerlinewidth=0.01pt,
    font=\footnotesize,
]{promptbox}
\makeatletter
\let\orig@promptbox=\promptbox
\def\promptbox{
  \@ifnextchar[{\promptbox@opt}{\orig@promptbox}
}
\def\promptbox@opt[#1]{
  \orig@promptbox[frametitle={#1}]
}
\makeatother
\DeclareRobustCommand{\DE}[3]{#3}
\DeclareRobustCommand{\VAN}[3]{#3}

\title{Awes, Laws, and Flaws From Today's LLM Research}

 \author{Adrian de Wynter \\
   Microsoft and the University of York \\
   \texttt{adewynter@microsoft.com} \\
}

\begin{document}
\maketitle

\begin{abstract}
    We perform a critical examination of the scientific methodology behind contemporary large language model (LLM) research. 
For this we assess over 2,000 research works released between 2020 and 2024 based on criteria typical of what is considered good research (e.g. presence of statistical tests and reproducibility), and cross-validate it with arguments that are at the centre of controversy (e.g., claims of emergent behaviour). 
We find multiple trends, such as declines in ethics disclaimers, a rise of LLMs as evaluators, 
and an increase on claims of LLM reasoning abilities without leveraging human evaluation. 
We note that conference checklists are effective at curtailing some of these issues, but balancing velocity and rigour in research cannot solely rely on these. 
We tie all these findings to findings from recent meta-reviews and extend recommendations on how to address what does, does not, and should work in LLM research. 

\end{abstract}

\section{Introduction}\label{sec:Introduction}

Large language models (LLMs)\footnote{We use the term `LLM' loosely to refer to generative text-to-text models with sizes of or above 1B parameters.}
 are a powerful technology. 
They can follow instructions and output coherent, persuasive text. 
This has made them the centre of attention in academia, industry, and the media. 
Given the potential funding associated with these technologies, it should not be a surprise that news and research works are sometimes accompanied by bold claims about their capabilities. 

It has been said that the focus of AI research is the approaches, rather than the results. 
That is, less importance is allocated to issues around experimental protocols when compared to other fields \cite{doi:10.1126/science.adf6369}. 
For example, papers sometimes lack sufficient details for independent verification \cite{gehrmann2023repairing,10.1145/3514094.3534196}; or report aggregate performances (e.g., accuracy) without providing detailed protocols or error breakdowns \cite{bouthillier2021accounting,gehrmann2023repairing,doi:10.1126/science.adf6369,10.1145/3571884.3604316,10.1145/3514094.3534196}. 
This is common in health sciences \cite{doi:10.1126/scitranslmed.abb1655}, security \cite{10.1145/3576915.3623130}, and recommender systems \cite{Cremonesi_Jannach_2021}--all areas where LLMs are increasingly being applied. 

We argue, however, that in LLM-related research there has been a shift towards \emph{result-driven experimentation}, partly due to their availability and capabilities; but also due to scrutiny from the media, funding sources, and the field. 
In theory, this should mean that the scientific community ought to allocate the same relevance to experimental protocols and good research practices as other fields. 

This is easier said than done, since LLMs may be closed source, expensive to train, and/or limited to a versioned API call. 
Solving a problem could be done one single prompt, and their ability to generate text reduces the time taken to write and produce papers. 
A combination of these could lead to a rapidly-increasing volume of scientific articles with strong claims and lacklustre experimental practices. 
However, LLMs as a technology are not the only factor: evaluating these models is notoriously difficult, researchers are under pressure to keep up, and peer-reviewing systems are usually overloaded. 
Still, all of this impedes the scientific community's ability to transparently evaluate and understand LLMs, and to ensure their responsible use. 
It also poses questions about the validity and trustworthiness of some findings and claims, especially around LLM capabilities. 

Our work is a statistically motivated critique, meant to \textit{systematically and critically examine} the scientific methodology employed in LLM research, and to \textit{quantify} the extent to which these issues occur in the literature.\footnote{Code and partial, anonymised data will be released in \url{https://github.com/adewynter/awes\_laws\_and\_flaws}} 
We look at what does, does not, and should work in LLM research, and, based on that, extend recommendations to the scientific community at large to retain velocity without sacrificing innovation or good research practices. 

\subsection{Contributions}
We evaluate 2000+ scientific articles that have an LLM as the focus of study, and based on the presence of a set of criteria. 
Most of these come from reproducibility checklists for premier conferences.\footnote{See, e.g. \citet{AAAI,NeurIPS} and \citet{ACLARR}.} 
The rest are chosen based on claims at the centre of controversy, such as using LLMs as evaluators, or assertions of emergent behaviour. %

Our analysis shows various trends, such as a decreased emphasis on ethics or appropriate research protocols (accounting for versioning, declaring call parameters, etc.), and a rise on the use of LLMs as evaluators. 
There is also an increase of works in non-English languages and a steady number of limitation disclaimers. 

The takeaways of this study, however, are that the field appears to be increasingly rushing on producing papers lacking rigour experimentally and ethically; while certain conference mechanisms, such as ACL's enforcement of a limitations section, appear to be effective at curtailing some of these trends. 
This is important, as plenty of papers rely on recency and claims of SOTA for novelty--namely, citing non-peer reviewed sources. 
While not a problem \textit{per se}, the robustness of the sources, and hence the paper's arguments and results, could be put into question if they have not been carefully validated or used reliable experimental protocols. Take, for example, claims of emergent behaviour in LLMs \cite{wei2022emergent,GPT3} and the fact that better statistical methods make them `evaporate' \cite{schaeffer2023are}.

It is difficult to maintain selectivity, novelty, and speed. 
We then suggest a middle ground where we educate ourselves and the community at large on what constitutes a paper with robust experimental practices. 
This middle ground comes as a series of recommendations to the field at large on how to maintain rigour in research, all without sacrificing velocity in research and innovation.

\section{Background}\label{sec:background}

Some known challenges to AI research are particularly salient in LLM literature, although might require reframing. 
In this section we describe and tie them to the criteria and focus of our analysis. 

\textbf{Reproducibility:} LLM-related reviews have found that the most downloaded models in Huggingface do not consistently provide the same amount of information in their documentation \cite{cardsreview}; 
and an analysis on the experimental protocols of over 40 chat-based LLMs  (e.g., availability of data, weights, licences, etc) found that, while many projects claimed some level of open-sourcing, documentation was `exceedingly rare' \cite{10.1145/3571884.3604316}. 
Beyond that, reproducibility itself is tied to stochasticity and versioning, which in turn raises concerns about the results themselves, particularly around the fact that errors and biases are rarely reported or analysed \cite{pang2025understandingllmificationchiunpacking}. 

One core aspect of reproducibility, beyond protocol declaration, is open-sourcing. 
For LLM research in particular, this doesn't readily apply to models behind APIs. 
While in the broader CS field open-sourcing has been trending upward, reproducibility remains difficult \cite{arvan-etal-2022-reproducibility-computational}. 
Further, in some areas it has been found to be no statistically significant difference in the presence of artifacts (code, data) when the mechanisms for reproducibility were implemented \cite{10.1145/3576915.3623130}. 
This in turn suggests that, due to LLM stochasticity and system limitations, open-sourcing alone is preferrable, albeit not a cure-all: but protocol declaration, however, remains important. 

\textbf{Measurement} is difficult in LLMs due to effectiveness and scalability of metrics. 
It has long been known that automated metrics like BLEU do not capture natural language generation well \cite{liu-etal-2016-evaluate,novikova-etal-2017-need}, and not correlate well with human judgements \cite{reiter-2018-structured,gehrmann2023repairing}, which themselves have their own complications \cite{clark-etal-2021-thats,van-der-lee-etal-2019-best}. 
While it has been argued that metric performance is not as important as tasks and methods, and mostly drive improvement within the task \cite{pramanick-etal-2023-diachronic}, benchmarks and leaderboards are the reigning way to define LLM success. 

These have problems scaling, however, given how flexible LLMs are at multiple tasks. 
Although there is a push to use them as evaluators, there is no consensus on the viability of this approach: arguments in favour \cite{chiang2024chatbotarenaopenplatform,chiang-lee-2023-large,liu-etal-2023-g,rethinkingsemantic,NEURIPS2023_91f18a12} are as plentiful as arguments against 
\cite{doddapaneni2024findingblindspotsevaluator,stureborg2024largelanguagemodelsinconsistent,chiang-lee-2023-closer,rtplx,rethinkingsemantic,LLMLXEval}. 
One reason why this approach is often called unreliable is due to the fact that LLMs might memorise their training data \cite{PlagiariseLee,dewynter2023evaluation}, including evaluation benchmarks \cite{sainz-etal-2023-nlp}; their results are very sensitive to the prompt's phrasing \cite{lu-etal-2022-fantastically,hida2024socialbiasevaluationlarge}; and that, generally speaking, an NLG system's performance, including evaluator systems, is strongly dependent on the choice of metric \cite{vondäniken2024measuredependenceautomatedmetrics,gao2025analyzingevaluatingcorrelationmeasures}. 
It is clear that carefully-chosen, diverse metrics and statistical tests are necessary for trustworthy results. 

\textbf{Claims:} Many aspects of LLM research are related to their (or lack thereof) capabilities. 
For example, it is often said that LLMs present emergent abilities. 
This is usually defined as their ability to solve more complex problems in a way not predictable by, say, parameter size \cite{wei2022emergent,GPT3}. 
This definition is slightly vague, given that the problems or their measures of complexity are not actually defined. 
Another common claim associated with LLM research are claims of artificial general intelligence (AGI). 
Analogous to emergence, AGI claims often rely on disparate definitions and goals \cite{blilihamelin2025stoptreatingaginorthstar}, which in turn makes results incomparable. 
Even the term `reasoning' is not always well-scoped (abductive? analogical? formal?; \citealt{huang-chang-2023-towards}). 
The relationship of these claims to rigorous experimental protocols are central to our work. Aside, recall from our earlier arguments that some of these claims fall apart under closer scrutiny, namely under more robust statistical tests. 

\textbf{Ethics and Inclusion:} LLMs are known to cause and propagate multiple harms, in addition to have very strong multilingual capabilities. 
It has then become an active focus area to ensure that LLMs are used in an ethical and inclusive manner for multiple audiences. 

Note, however, that ethical concerns and evaluations are \textit{system-dependent}: disclosures, protocols, and the concerns themselves are not universally tied to English-based language modelling, or even NLP. 
For example, alignment of models is usually carried out within a universal value system (e.g., `always be honest'), but this has been called out for being pragmatically inadequate \cite{varshney2025scopesalignment}. 
Likewise, evaluations under a single prompt tend to fail when working with culture-specific tasks \cite{,cheng2025englishevaluatingautomatedmeasurement}. 

Add to that the fact that the risks called out by researchers may not be the same risks considered by laypeople \cite{karamolegkou2024ethicalconcernidentificationnlp}, 
or even other fields: only 2\% of non-CS papers using LLMs call out ethical considerations \cite{transforminglandscapes}, perhaps due to less concern or familiarity with this area. 
Note that the focus of this work is \textit{LLM-centred research}, so papers from fields other than NLP are also part of our analysis. 

Hence, authors should have a responsibility to educate others on the impact of their technology, and be aware of what matters for broader audiences. 
Therefore, ethical and inclusive considerations %
should remain constantly present as part of \textit{any} LLM-related work.

\textbf{Recommendations:} Studies and meta-reviews of AI often recommend good practices (e.g., reporting carbon footprints \cite{ReportingFootprints}; reproducibility checklists and experimental protocols \cite{reproducibilityml,Gundersen_Kjensmo_2018,VANDERLEE2021101151}; self-contained artifacts \cite{arvan-etal-2022-reproducibility-computational}; and metadata for corpora \cite{gebru2021datasheets}. 
These suggestions are often not heeded: \citet{gehrmann2023repairing} noted that, out of 66 articles from leading conferences, only between 15\% and 35\% of the recommendations were partially followed. 
The practices themselves might not even be sufficiently impactful \cite{10.1145/3576915.3623130}, given that other experimental protocols (e.g., sampling, initialisation, hyperparameters) may also impact reproducibility  \cite{bouthillier2021accounting}. 
In this work we are unable to directly address this lack of impact, although we do note in \secref{recommendations} where it is possible to have better momentum without (completely) enforcing checklists in conferences.

\section{Methods}\label{sec:methods}

\subsection{Corpus}

Our corpus is comprised of works that cited the peer-reviewed GPT-3 paper \cite{GPT3} and the GPT-4 technical report \cite{GPT4}. 
We make the assumption that the majority of the LLM literature references either of these articles. 
We examine this assumption more closely in \secstworef{conclusion}{limitations}; and evaluate its longevity in a follow-up study done a year after our data cutoff (\appref{googlescholar}). 

We retrieved the top 1,000 papers sorted by citation numbers for both articles in Google Scholar\footnote{\url{https://scholar.google.com/}} with Publish or Perish \cite{PoP}; 
and the top 2,000 papers by citation number for GPT-3 in Scopus.\footnote{\url{https://www.scopus.com/}} 
The disparity is due to Google Scholar indexing peer-reviewed works and preprints, and Scopus only indexing peer-reviewed articles. 
At the time of writing this paper, the GPT-4 technical report has yet to be peer-reviewed. 
All queries were ran for papers published or released up to 10 June 2024. 
Considerations around the representativeness of this corpus are in \secref{limitations}. 

We used the \arxiv{} API\footnote{\url{https://info.arxiv.org/help/api/index.html}} to retrieve the full paper, and parsed either the source into text. 
The final, deduplicated, unlabelled corpus is $3,914$ texts. 

\subsection{Evaluation Criteria}

We labelled our corpus based on a set of criteria (labels), categorised in four groups: Research Features, Structural Features, Arguments Made, and Indicators. 
Groups and labels are in \tabref{all_criteria}, with specific definitions--as prompts--in \appref{prompt}. 

Research Features, Structural Features, and Arguments Made are the core evaluation criteria used in the rest of our paper. 
Indicators is a filter for our corpus: we were only interested in research articles with an LLM as the subject of research.

\begin{table*}[ht]
    \centering
    \begin{tabular}{|l|l|}
        \hline
        \textbf{Research Features} & \textbf{Arguments Made} \\ \hline
        $^*$Presence of statistical significance tests & $^\dagger$Claims of SOTA results \\
        $^*$Declaration of model versions (or API) & $^\dagger$Claims that the model can reason \\
        $^*$Declaration of parameters for calls made &  $^\dagger$Claims that the model cannot reason \\        
        $^*$Accounting for stochasticity of the calls & $^\dagger$Claims of emergent behaviour \\
        $^x$Evaluation of non-English languages and/or dialects &  $^\dagger$Claims of super-human intelligence \\
        $^\dagger$Use of human, automatic, or LLM-based evaluators &  \\ \hline
        \textbf{Structural Features} & \textbf{Indicators} \\ \hline
        $^*$Presence of a limitations section & LLM is the subject of the research \\ 
        $^*$Presence of an ethics section & Type of text (research, book, or opinion) \\ 
        $^x$Presence of error breakdowns &  \\ 
        Presence of negative results &  \\ \hline
    \end{tabular}
    \caption{Criteria for our analysis. Research Features, Structural Features, and Arguments Made are the core subject of our work. Most labels are binary labels (yes/no); but Research Features also include `na' (not applicable), and evaluators is a set (\{human, automatic, LLM, na\}). 
    Indicators is a filter to only select LLM-centred work research articles. 
    We use ($^*$) for criteria from conference checklists and ($^x$) for those recommended--but not implemented--as good research practices, and ($^\dagger$) for claims requiring closer examination. 
    See \appref{prompt} for definitions.}
    \label{tab:all_criteria}
\end{table*}

\subsection{Labelling}
Given the large volume of data and budget constraints, we were unable to perform a full human-based labelling work. Instead, we labelled the data with GPT-4 omni (version: gpt4-o-2024-05-13). 
To ensure reproducibility, we set the temperature to zero, the maximum output tokens to 256, and left other parameters as default. 
To improve accuracy we split in batches our labelling calls, totalling five different prompts \cite{LLMLXEval}. 
All our calls were done through the Azure OpenAI API and the analysis done with a consumer-grade laptop. 

To ensure trustworthiness of our results, we measured the model's reliability by sampling 100 papers per criterion, and manually labelling them. 
We found the model's results to be reliable to an average 91.91 $\pm$ 1.22\% accuracy, with a 95\% confidence interval. 
This number varies broadly across criteria. 
A full breakdown of reliability and analysis of performance is in \appref{reliability}. 

Regarding the label set, the model was instructed to return binary labels (\{y, n\}) for all criteria, except for most Research Features labels, which also included a relevance label (\{na\}). 
To determine the evaluators used in a given paper we used the set \{human, LLM, automatic, na\}; 
and for the type of text, \{\text{book}, \text{article}, \text{opinion}\}. 
For our topic analysis, we requested a primary subject for the paper in a few words, and then manually clustered them.
Prior to our experiments, including topic analysis, we filtered the corpus by selecting all research articles with an LLM as the main subject of study. 
The final size of the data was $2,054$ papers.

\section{A Review of LLM-Centred Literature}\label{sec:results}

We provide four analyses: corpus composition (\secref{composition}), composition over time (\secref{yearly}), the relationship between citations and criteria (\secref{citations}), and yearly trends on relationships between citations and criteria (\secref{betterorworse}). 
The full breakdown of results is in \appref{extended}. 
Throughout this section, we use \emph{relevant papers} to refer to these that did not score `na' in the criterion discussed.

\subsection{How Many Papers Did What?}\label{sec:composition}

\begin{figure}[h]
    \centering
    \includegraphics[width = \linewidth]{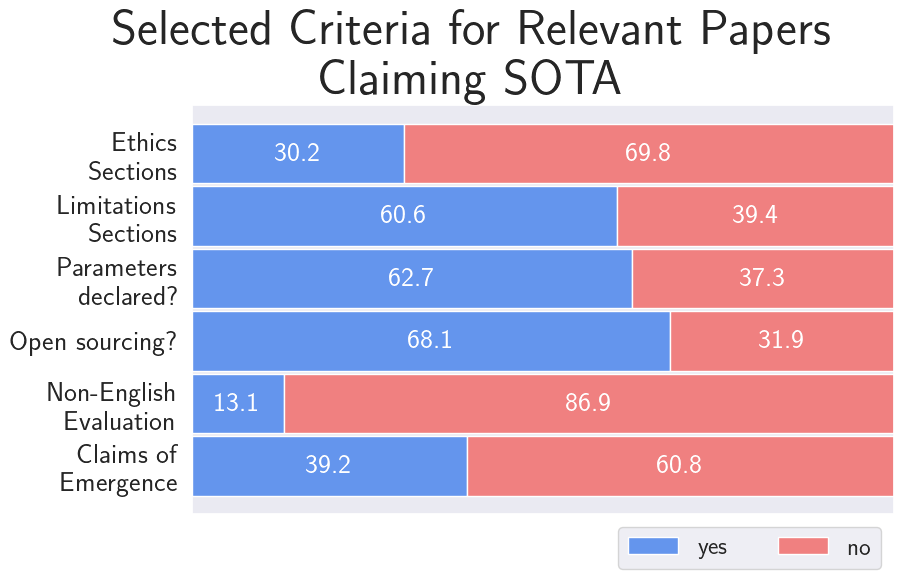}
    \caption{Breakdown of the corpus for selected criteria, narrowed down by papers claiming SOTA results. 
    While open-sourcing, declaration of experimental protocols, and limitations are relatively high (60\%+); ethics sections (30\%) and evaluations in languages other than English (13\%) are comparatively low.}
    \label{fig:papers_aggregate}
\end{figure}

We found that 57\% of the articles claimed SOTA results. 
A third of these contained or addressed ethical considerations related to their research; 
13\% performed evaluations in languages other than English; 
and 39\% did not include limitation sections about their experimentation (\figref{papers_aggregate}). 
Only a quarter of them included statistical tests to support their claims--close to the 23\% found by \citet{VANDERLEE2021101151}. 
This is lower than for papers that do \emph{not} claim SOTA. 
Further results are in \appref{corpuscomposition}. 

\begin{figure*}[h]
    \centering
    \includegraphics[width=\linewidth]{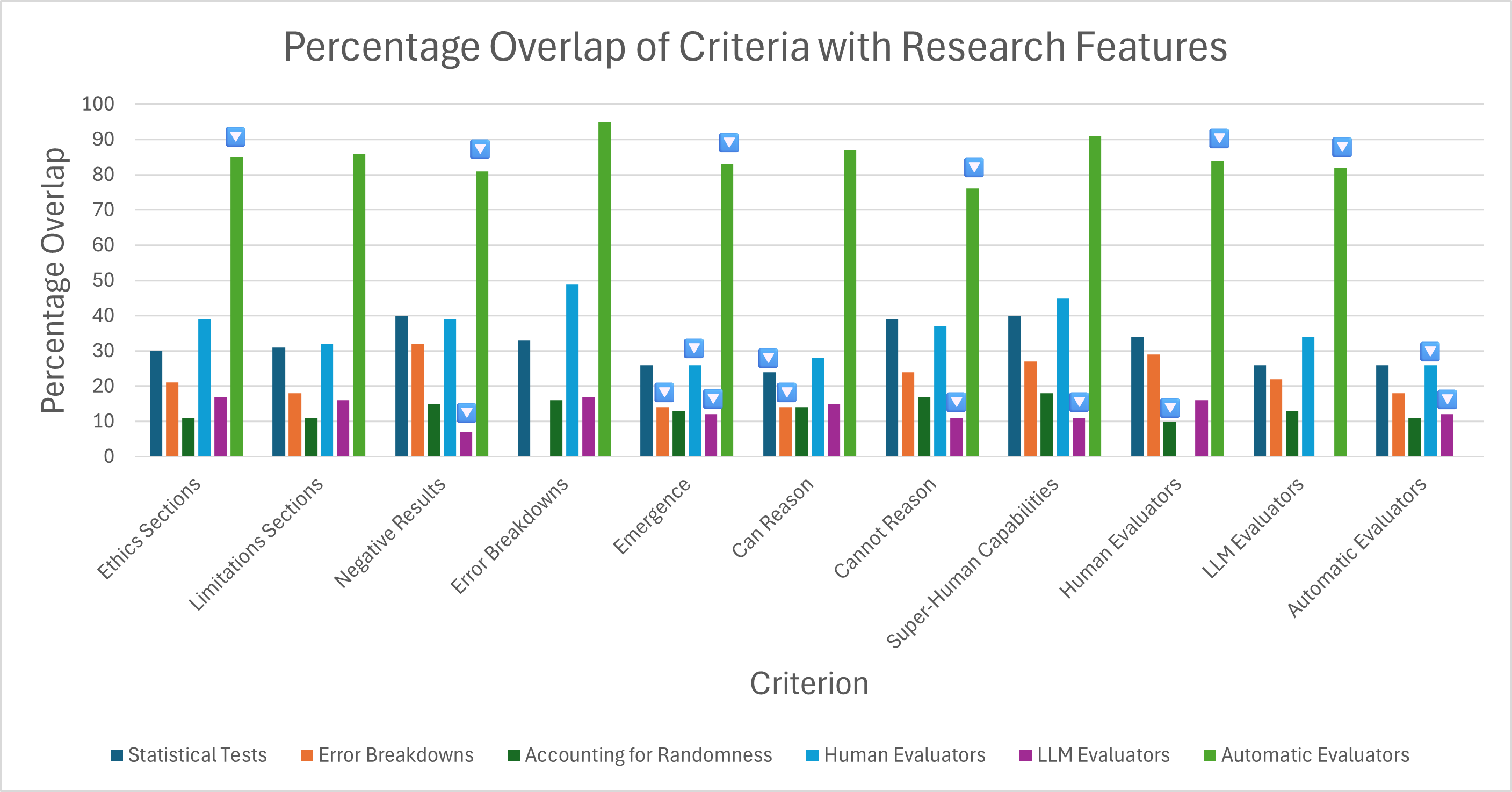}
    \caption{Percentage overlap for selected criteria with respect to Research Features. The claims whose percentage is below the average for papers claiming SOTA are marked with a down arrow. 
    }\label{fig:proportionalresults}
\end{figure*}

In terms of criteria overlap (\figref{proportionalresults}), from the papers claiming SOTA and emergent capabilities, only a quarter of them relied on statistical tests or had error breakdowns. 
The articles usually included automatic metrics, and reliance on LLM evaluators alone was exceedingly rare. 
Claims of LLM reasoning capabilities were often done with LLM evaluators and not human evaluators; conversely, claims that they cannot reason were done with human evaluation alone.

Compared to the entire corpus, SOTA papers use fewer measures of statistical significance overall, and the use of automated metrics is more common than other types of measurements. 
That said, SOTA papers \textit{only} using LLMs as evaluators were rare. 
Many relevant SOTA papers did report model versioning (73\%), and open-sourced their work (68\%), a number slightly higher than the one found by \citet{arvan-etal-2022-reproducibility-computational}, indicating growth. 
A full breakdown of the results is in \appref{composition}.

\subsection{What Changed Over Time?}\label{sec:yearly}

In our corpus, 46\% of the papers belonged to the first half of 2024, contrasting with 22\% and 25\% for 2022 and 2023. 
Given that our 2024 subset only comprised half of a year--but twice as many works as in 2022 or 2023--we may assume that this analysis will not incur any recency bias. 
\begin{figure}[h]
    \centering
    \includegraphics[width =\linewidth]{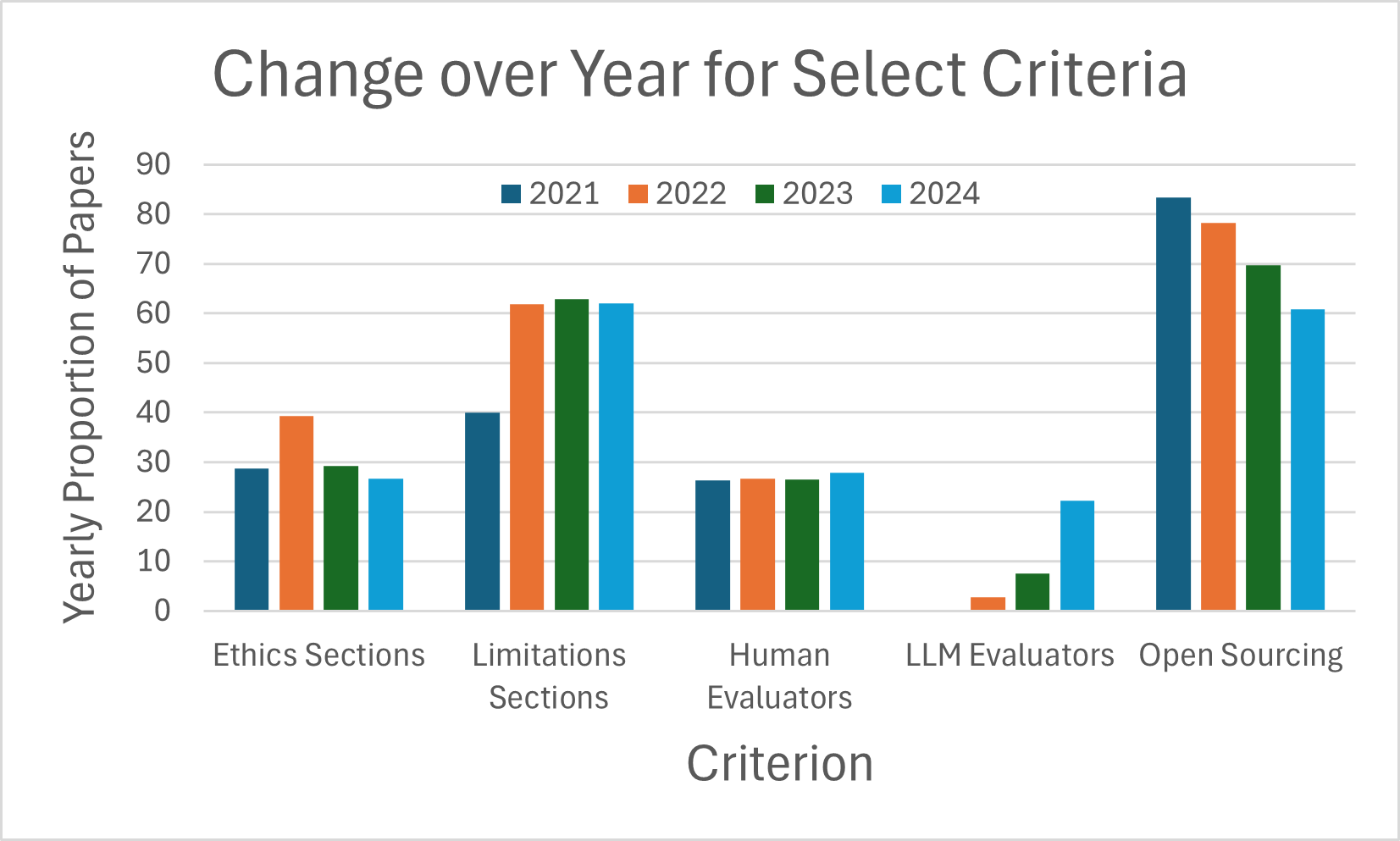}
    \caption{Change over time for selected criteria. Between 2022 and 2024 there was an increase in the use of LLMs as evaluators; and declines on the presence of ethics disclaimers and open-sourcing. The use of human evaluators and limitation sections are steady.}
    \label{fig:over_time_general}
\end{figure}

For this analysis we worked with relevant papers claiming SOTA from 2021 onwards. 
Between 2023-2024 we observed declines in the absolute percentage of several criteria, such as the presence of ethics disclaimers, open-sourcing, claims of emergence, and statistical tests (\figref{over_time_general}). 
There was an increase in the claims that these models can reason (15\%); and a decrease in the claims that they cannot. 
Presence of limitation sections, error breakdowns, dialect evaluation, and relying on human evaluation remained steady ($\pm$1\%). 
The use of LLM evaluators underwent an uptick (15\%), but these papers generally have lower-than average proportions of research protocols. 
In terms of topics, papers have seen a doubling in subjects like multimodality and safety and security. 
Full results for topic changes are in \appref{yearlytopics}. %

\subsection{Do Papers With Certain Criteria Get More Citations?}\label{sec:citations}

We reviewed the relationship between the presence of criteria on SOTA papers and the number of citations they received. 
Given that the corpus is long-tailed, we limited our analysis to the top 1,059 relevant papers, which contain 91\% of all citations. 

We split our corpus in two (texts with and without a given criterion), and did a two-sample Kolmogorov-Smirnov test to determine the probability that both samples came from the same distribution. 
In this test, a high probability implies \textit{no significant difference} between the samples. 
In this case we may then conclude that the presence of the criterion does not impact the number of citations, as they likely are drawn from the same distribution. 
Conversely, a low probability of being drawn from the same distribution allow us to conclude that the criterion is \textit{not} related to the number of citations. 

For our corpus we first calibrated the $p$-value to $<0.05$ for all criteria, which means that we expect to be wrong about our conclusions 5\% of the time. If the test's $p$-value is below our calibration threshold, we may reject the null hypothesis $H_{0}$ that the samples are related (i.e., that having one has an impact on the other). 
See \appref{kolmogorovsmirnov} for a detailed explanation of this test and full results.

\begin{figure}[h]
    \centering
    \includegraphics[width =\linewidth]{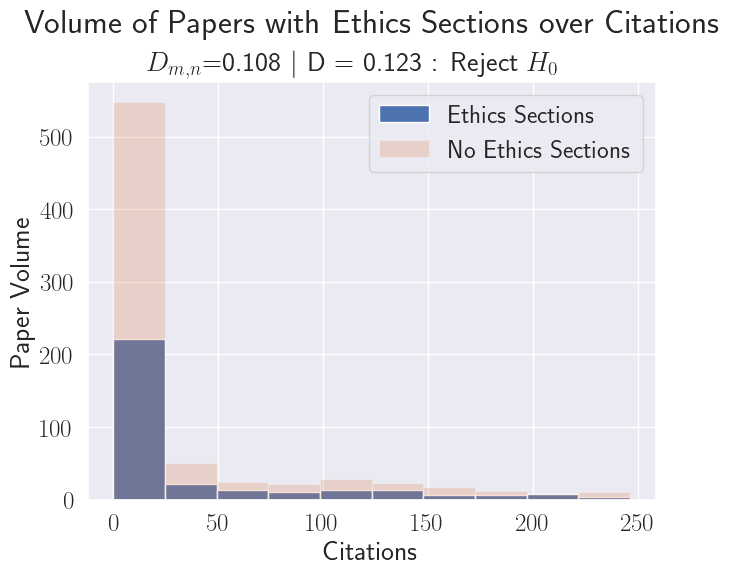}
    \caption{
        Results of a Kolmogorov-Smirnov test on papers with (blue) and without (orange) ethics sections. 
        The $p$-value for the test is p=$0.016$. 
        This indicates that, if both samples were drawn from the same distribution, the probability that they are as far apart as observed
        is 1.6\%. 
        Since they \emph{were} drawn from the same distribution, we may conclude that the presence of ethics sections \emph{does impact} the number of citations received. 
    }
    \label{fig:ks_ethics}
\end{figure}

From our experiments, we rejected $H_{0}$ in the presence of ethics (\figref{ks_ethics}) and limitation sections; the use of LLM and automatic evaluators; open-sourcing; and claims of reasoning. 
This means that the presence of these on a paper had an impact on the citations it received. 
For other criteria (error breakdowns, evaluation of non-English languages, claims of emergence, and negative results) we were unable to reject $H_{0}$. Hence, their presence did not affect the citations received.

\subsection{Are We Getting Better or Worse?}\label{sec:betterorworse}

We analysed the relationship between citations and the presence of our criteria as yearly trends from 2021 onwards. 
We evaluated this relationship as the \emph{gap} (absolute percentage difference) between citations of papers containing the criterion and these that do not, aggregated by year, and measured the \emph{change} in this gap: positive changes imply that papers containing the criterion are cited more often. 
See \appref{yearlychanges} for examples and the full results. 

We observed an increase in the gap for ethics sections in 2022-2023 (+10\%), but a noticeable drop in 2023-2024 (-50\%). %
Other criteria had similar patterns, such as presence of statistical tests, limitation sections, and negative results. 
Note that 2024 gap drops are expected, given the recency of these papers. 
There were also upticks: LLMs as evaluators and non-English evaluations both had increases in citations.
Unlike \secref{citations}, recency bias is a concern in this experiment, which we discuss in the next section.

\section{Discussion}

\subsection{Volume Analyses}
Few papers claiming SOTA addressed ethical considerations, and this number is in decline. 
The proportions of statistical tests and open-sourcing in these papers are in line with critical literature for other CS fields. 
We also found open-sourcing and research protocols to be declining. 
Likewise, statistical tests are less often used than in LLM-related literature \textit{not} claiming SOTA. 
Given the considerable volume of research contributed by the first half of 2024, these trends suggest rising rushed research and lack of rigour. 

The use of LLMs as evaluators increased significantly, and most papers using them claimed SOTA results. 
This increase could be explained by technological developments (GPT-4, the standard of LLM-based evaluation, was released in 2023) and replication of experimental work. 
These works often coupled LLMs with other evaluator classes (e.g., humans), but eschewed measures of statistical significance or accounted for randomness in the calls. 
This also suggests rushed research, but at a lesser extent given the presence of other evaluators. 

Steady criteria, namely limitation sections, may be explained by being a requirement of *ACL conferences \cite{ACLARR} since 2022: indeed, there is a $\sim$40\% increase of papers with this criterion in 2021-2022. %
The increased focuses on multimodality and safety indicate that this technology has matured beyond research; or, at least, that it is being widely adopted beyond NLP. 

\subsection{Claims}
Claims of emergent behaviour were common--although in decline--but coupling them with statistical tests or error breakdowns were rare. %
On the other hand, claims that these models can reason were evaluated with LLMs and not humans; versus claims that they could not reason, typically done with human evaluation alone. 
Both suggest poor research practices: as per \secref{background}, metrics and evaluators are system-dependent, and a more robust approach would use more than one of both.

\subsection{Citations and Criteria}
We observed a statistically significant difference in the citations received by a paper when presenting ethics, limitations, and LLM evaluators. 
The first two may be explained by papers written for, or accepted at, a specific venue, and hence perhaps having higher quality. 
The latter, given its novelty, could be simply due to stronger claims. 
There was a statistically significant difference in the likelihood for a paper to be cited if it had an ethics section present or used an LLM as an evaluator. 
Due to likely recency bias, we were unable to conclude if it was because of a higher quality bar for the paper, the venue, or because of the claims made. 
It could be argued that ranking by citations induces a recency bias \cite{citationsimpact}; however, the (volume-based) results relying on citations as proxy do not change. Citations are a common metric to determine a paper's success. There is also considerable skewness on their distribution, with 91\% of the citations held by 25\% of the corpus. 
Still, we caution on the interpretation of citation-based time-wise results. 

\subsection{Notes and Alternative Explanations}
We observed decreases in citation rates in 2023-2024 for some criteria. This could have a simple explanation: newer papers have fewer citations. Hence we do not factor in these results in our conclusions. 
An uptick is \textit{not} expected, however. %

An alternative explanation for the decrease in ethics sections could be the multidisciplinary adoption of LLMs and the distinct ethical requirements for other fields. 
We consider matter-of-factly that ethics disclaimers in LLM-centred research have decreased, regardless of field. However, we do factor this observation into our recommendations, and discuss their relationships to venue requirements in \secstworef{conclusion}{limitations}.

\section{Conclusion and Takeaways}\label{sec:conclusion}
Our study is a critique of the methodology employed in LLM research. While statistically motivated, the centrepoints of our argument are related to taking stock of where we are and where are we going as a research community. 

We noted a yearly decline on the proportion of papers having ethics disclaimers, open-sourcing, and statistical tests. %
Articles claiming SOTA relied more on LLMs as evaluators, and had fewer statistical tests, that the ones that did not. 
They also reported less research protocols, especially in 2024. 
We consider these evidence of mounting overreliance, lack of rigour, and/or rushed research, especially with respect to measurements. 

That said, there were two major positive findings in our work. 
The first was a steady proportion of limitation sections, which we ascribed to requirements from conferences. 
The second was an increasing number of evaluations in non-English languages, and a rising interest on safety and security as a research subject. 
From both we conclude that a combination of available, powerful, LLMs; along with publication checklists \textit{does} have a net positive effect on the literature, and on a more diverse and inclusive research community.

We also found that LLM-centred research is rapidly increasing in volume, and consequentially it could be that our assumption that most papers will cite GPT-4 or GPT-3 does not necessarily hold in the future. 
To determine this, a year after the data cutoff for our paper we attempted to perform a follow-up study. 
This could have determined any rates of change and recency biases induced by our measurements, along with the validation, or discrediting, of our core assumption. 
However, we were able to only validate the latter (\appref{googlescholar}), because the tools we used in our work were no longer able to access Google Scholar. 
We argue that this is a concerning trend: the ability of surveys like ours to critically examine the field strongly rely heavily on open tools and publicly-available APIs. 
Losing that access could have ramifications on the ability of the field to perform self-examination. 

Nonetheless, our work underscores the need for more self-scrutiny and rigour by and from the field. 
This is not easy given the overload of authors and peer-reviewers. 
It is also not feasible to wait months for the (purported) `next big thing' to be peer reviewed if it is readily available as a preprint. 
While it could be said that critical reading is crucial, it would be na\"ive, however, to assume that this will be consistently done by all readers--e.g., laypeople, scientists not tasked with their peer review, etc. 
Our recommendations are designed to address this. 

To close, \citet{Cremonesi_Jannach_2021} mentioned tongue-in-cheek that in the context of recommender systems there was no reproducibility crisis, as in a crisis researchers reflect upon and revise their methodologies. 
Instead, they stagnanted due to overfocusing on the same subjects without any introspection. 
It is our hope that our findings encourage LLM researchers to reflect on how to push the field forward, while also carrying out research that is ethical, systematic, and open to criticism.

\section{Recommendations}\label{sec:recommendations}
Based on our observations from the previous section, we extend three recommendations. 
These call for specific features in papers that should be scrutinised during peer review. 
This is because, when researchers know that either the venue will enforce some features (e.g., limitation sections), or, that reviewers will ask for them (e.g., releasing artifacts), they will usually add them to the pre-review preprints. 
While this is not common outside of CS, adding these features within NLP will encourage researchers to reflect upon their work \textit{and} allow other readers to readily comprehend the scope of the research. 
Hence, we summarise our recommendations to the scientific community in three areas: impact analysis, measurement rigour, and transparency. 

\textbf{Impact Analysis:} Venues should continue (or begin) to enforce (short) sections that allow for easier critical evaluation of the work, and encourage self-reflexion and thoughtful research by the authors. 
The two sections are (1) limitations/scope, and (2) considerations on the broader impact of the work. 
The first must disclose the experimentation's boundaries and areas of improvement/future work. 
The second is not an ethics disclaimer, but still allows readers to understand the implications of the research. 
Peer-review feedback after reviews should be added in: as third parties without any stakes in the work, they are best positioned to provide informed critiques that other readers may miss, even if available elsewhere (e.g., OpenResearch). 
Neither section should count towards the page limit.

\textbf{Measurement Rigour:} Reviewer checklists should explicitly account for (but not require) statistical tests,\footnote{Their definition must also be included in the work.} number and type of metrics and languages evaluated, and classes of evaluators used. %
We do not call for their enforcement as that these are context-dependent. 
That said, since LLM work is primarily empirical and comparison-based, they are likely going to be required often. 

The idea is that measurement rigour should explicitly be part of an even evaluation of a paper's merits.
It could also allay some of the concerns from \secref{background} on system-dependence and reliability, and propose areas for further work. 
This \textit{does not} mean that papers using LLM evaluators alone should be discounted outright, but their methodologies should be scrutinised closely. This is especially important for works using a single prompt or LLM evaluator, such as this one. 
Without a meta-evaluation, the results should be deemed unreliable.

\textbf{Transparency} in LLM research is tricky, but not unattainable. 
Terminology (e.g., AGI, emergence, reasoning) must be formally and carefully defined if evaluated in the work. 
Declarations of prompts, call parameters, and versioning must be enforced. 
Reviewers should be encouraged to seek an error analysis section focused on the LLMs' responses: LLMs are notoriously unreliable, and they are better understood when analysing their responses qualitatively. 
On open-sourcing, it is worth noting that a model's weights being open-sourced does \textit{not} constitute transparency, if neither the code nor the data were released. 
Unlike earlier recommendations, this is a matter of semantics and may be caught during peer review.  

At a more meta level--and not strictly related to peer-review--our finding that public APIs could not access certain sites signifies a loss of transparency that could have repercussions on the ability of the field to perform self-examination. 
We recommend that these APIs should have their access re-enabled.

\section{Ethical Considerations}
Open data is crucial for good research, but ethical and licencing considerations limit us from releasing the corpus with texts and personally-identifiable information. 
We release the code for our analysis under a permissive licence (MIT), and the annotated, anonymised data without texts. 
To avoid overloading the services we rate-limited our requests, in compliance with their terms of use; and the crawling code will not be released. 

\section{Limitations}\label{sec:limitations}

\subsection{Reliability of Automated Labelling} 
The community remains divided on the feasibility of using LLMs as evaluators.
We argue that the reliability and conclusions drawn from using this technology vary with the problem and experimentation protocols used. 
We mitigate potential concerns by evaluating the performance of the model with statistical tests: our analysis showed that GPT-4's confidence bounds and accuracies were reliable.

\subsection{Corpus Representativeness}
Our analysis is limited to works available on Google Scholar and Scopus citing the GPT-3 and GPT-4 papers. 
This might not represent the entire body of research on LLMs. 
As the literature and the technology evolves, we expect this assumption to hold less weight. 
However, as it stands, a year after the cutoff data for our paper, both papers are still very dominant in the literature. 
See \appref{googlescholar} for a follow-up study evaluating our assumption. 

This also in turn overlooks a potential issue with citation-based ranking: well-known authors, venues, and institutions could have more citations just by virtue of being known. 
They could also evolve over time, inducing sampling biases \cite{citationsimpact}. 
This is, however, a common shortcoming to papers like ours (c.f., \citealt{pramanick-etal-2023-diachronic}), and, as argued, not a detractor from the core points of our work.

\subsection{Venues in Scope}
One of the main findings of this work was that the requirements from some venues (e.g., ACL) around mandatory sections were successful at maintaining their presence stable. 
However, a careful study could distinguish between venues to ablate out correlations. Nonetheless, the experimental setup, relying on open APIs, means that a large volume of papers evaluated may not necessarily be accepted, or even submitted, to these venues. This makes said experimentation tricky and likely the subject of future work.

\subsection{Criteria Established}
Due to time constraints we were unable to address an increasingly problematic issue in LLM research: synthetic data and use of possibly-contaminated benchmarks. 
We leave that exploration for future work. 

\subsection{Quality Assessment}
Our study focused on evaluating the presence of the criteria, as opposed to assessing the quality of the research methodologies employed or the arguments made. 
This suggests that, although we have observed a decrease in certain metrics, citations could still remain skewed to well-argued papers. 
That said, automated measure of argument quality is subjective, multi-faceted, and requires a good grasp on the pragmatic context (mostly historical trends, in this case). 
We leave this for future work.

\section*{Acknowledgements}

The author wishes to thank arXiv for use of its open access interoperability; A. Jangra and S. Visser for comments on an early version of this work; and the anonymous reviewers, whose informed critiques were fundamental to improving this work and were incorporated into the final version. 

\DeclareRobustCommand{\DE}[3]{#2}
\DeclareRobustCommand{\VAN}[3]{#2}
\bibliography{biblio}

\clearpage
\appendix
\section*{Appendix}
\section{Prompts}\label{app:prompt}

We split the labelling process to simplify the calls and because new criteria were added as the experimentation progressed. They can be found in \promptstworef{prompt2}{prompt3} (criteria) and \promptref{topicprompt} (topic analysis). 
We used a single exemplar for each prompt. They were hand-picked from a paper that was not present in the corpus, and manually tuned for accuracy on a subset of the data ($n=10$) before labelling the full corpus. 
We requested the model to output a string during labelling. This string would be the rationale (for `na' labels), and the verbatim matching line of the paper otherwise. 
This technique has been shown to improve the model's performance to out-of-distribution entries \cite{Brahman_Shwartz_Rudinger_Choi_2021}, and was helpful on analysing the performance of the model, which may be found in \appref{reliability}. 

\captionsetup[table]{name=Prompt}
\setcounter{table}{0}

\begin{table*}[h]
    \begin{tabular}{|p{0.95\linewidth}|}
    \hline
\cellcolor{gray!5}I am going to link a scientific paper. Tell me if the paper contains: \\
\cellcolor{blue!15}- Claims of emergence \\
\cellcolor{blue!15}- A LLM (GPT-4, Gemini, etc) or SLM (Llama, Phi, etc) as the main subject of study \\
\cellcolor{blue!15}Additionally, tell me the type of paper it is. It can only be one of \{research, book, opinion\}. \\
\cellcolor{blue!15}research papers contain experiments; opinion pieces are subjective; and books collect and survey results. \\
\cellcolor{red!15}- Statistical significance tests (Pearson correlation, Welch's t-test, etcetera): NOTE: they must be clearly indicated. \\
\cellcolor{red!15}- Claims of new state-of-the-art (SOTA) results \\
\cellcolor{red!15}- Claims that the model can reason \\
\cellcolor{red!15}- Claims that the model CANNOT reason \\
\cellcolor{red!15}- Claims of super-human intelligence \\
\cellcolor{red!15}- Limitations section \\
\cellcolor{red!15}- Ethics section \\
\cellcolor{red!15}- Negative results \\
\cellcolor{gray!5}Answer everything with "y" or "n", and add an explanation separated by a pipe. If you pick "y", return the verbatim first line matching. \\
\cellcolor{gray!5}For example, \\
\cellcolor{gray!5}<EXEMPLAR GOES HERE> \\
\cellcolor{gray!5}|begin paper| \\
\cellcolor{gray!5}<TEXT GOES HERE> \\
\cellcolor{gray!5}|end paper| \\\hline
\end{tabular}
\caption{Labelling prompt for the Indicators and parts of Arguments Made (in blue) and Structural Features and the remaining Arguments Made (in red). 
Other areas are shared between both prompts; but blue lines do not appear in red, and viceversa. 
Indicators had $99\%$ accuracy. 
LLM-as-a-subject and emergent behaviour had lower accuracy and looser confidence intervals ($89.0 \pm 6.8$ and $83.0 \pm 8.1$, respectively). 
Inspecting the output showed that the main cause of failure was GPT-4o returning (leaking) the exemplar for that criterion and ignored the input. 
The red prompt had good accuracy (between $93 - 100\%$) with tight confidence intervals ($\pm 0.0 - 5.5$).}
    \label{pro:prompt2}
\end{table*}

\begin{table*}[h]
    \begin{tabular}{|p{0.95\linewidth}|}
    \hline
\cellcolor{gray!5}I am going to link a scientific paper. Tell me if the paper contains: \\
\cellcolor{blue!15}- Versions of the LLM tested \\
\cellcolor{blue!15}- Parameters of any calls done to the LLM \\
\cellcolor{blue!15}- Accounting for randomness of the LLM \\
\cellcolor{blue!15}- Open sourcing of the data \\
\cellcolor{red!15}- Error breakdown analysis (breakdown per-classes for its performance) \\
\cellcolor{red!15}- Evaluation of languages other than English \\
\cellcolor{red!15}- Evaluation of dialects, and not just the main language \\
\cellcolor{gray!5}Answer everything with "y" or "n", or "na", and add an explanation separated by a pipe. If you pick "y", return the verbatim first line matching. \\
\cellcolor{gray!5}You should only use "na" if the criterion is not relevant (for example, if the paper does not produce a dataset, opensourcing should be "na"; or if there are no calls to LLMs all criteria should be "na"). \\
\cellcolor{red!15}Additionally, tell me what type of evaluation metrics the paper uses: automatic, human, LLM, na. Give your response as an array (e.g., [LLM, automatic]) and provide lines verbatim with a pipe for all. \\
\cellcolor{red!15}Use automatic for BLEU, ROUGE, BERTScore, etc. LLM if they use an LLM (GPT-4, e.g.) or SLM (Llama) for labelling. Only return "na" if there is no evaluation performed. \\
\cellcolor{gray!5}For example, \\
\cellcolor{gray!5}<EXEMPLAR GOES HERE> \\
\cellcolor{gray!5}|begin paper| \\
\cellcolor{gray!5}<TEXT GOES HERE> \\
\cellcolor{gray!5}|end paper| \\\hline
\end{tabular}
\caption{Labelling prompt for the first (blue) and second (red) parts of Research Features. 
The model had low performance on open-sourcing (74\%), and acceptable (though low) accuracy in the other criteria. 
Non-English and dialects had good accuracy ($98$ and $100\%$). 
Examining GPT-4o's reasoning for open-sourcing showed that it sometimes interpreted this label as only applicable if related to an LLM. 
The type of evaluator had varying performances even though it was the same label: failures in human and LLM evaluators were mostly related to missing, rather than mislabelled, entries. 
    }
    \label{pro:prompt3}
\end{table*}

\begin{table*}[h]
    \begin{tabular}{|p{0.95\linewidth}|}
    \hline
\cellcolor{gray!5}I am going to link a scientific paper. Tell me what is the primary and secondary subject of the paper, in at most three words each. \\
\cellcolor{gray!5}Return the extract of the paper, verbatim, that correspond to why you picked that subject.\\
\cellcolor{gray!5}For example,\\
\cellcolor{gray!5}<EXEMPLAR GOES HERE> \\
\cellcolor{gray!5}|begin paper| \\
\cellcolor{gray!5}<TEXT GOES HERE> \\
\cellcolor{gray!5}|end paper| \\
\cellcolor{gray!5}- primary:  \\ \hline
\end{tabular}
\caption{Topic clustering prompt. The output distribution was very wide, with about 1,000 lexically unique entries. Manual verification was needed for most topics to reduce and classify it to our 10 primary topics.}
    \label{pro:topicprompt}
\end{table*}

\section{Labeller Reliability}\label{app:reliability}

The reliability (accuracy within a confidence interval) of each of the criteria is in \tabref{cieval}. 
Our core assumption is that the distribution of labels is normal. 
We then calculated the accuracy of our annotator (technically, the prompt) to within a 95\% confidence interval (CI) with a Student's t-Test. 
 sampling i.i.d. $n\approx100$ papers and manually labelling them. 
Note that for the choice of $n$ amounts to approximately 5\% of the relevant data. 
Some papers did not contain the criteria we evaluated (e.g., the LLM-as-an-evaluator metric is rare in papers prior to 2023), or were too skewed (dialect evaluations are very scarce); 
so we sampled extra and only evaluated the relevant criterion. 

Overall, the model (prompt) performs well as a labeller, although certain criteria were certainly better-performing than others (e.g. open-sourcing versus SOTA claims). 
We partially attribute this to prompting. 
However, a closer inspection of the papers and the model's rationale noted that the model tended to overlook content, sometimes verbatim, matching the criteria. 
The model had a tendency to make a liberal interpretation of the prompt: for example, for open sourcing, the model sometimes indicated that no open sourcing was performed because no LLMs were tested (which was not specified in the instructions; see \promptref{prompt3}). 
It also tended to frequently inject content about downloading the film `The Nun II' from Reddit. 

\captionsetup[table]{name=Table}
\setcounter{table}{3}

\begin{center}
\begin{table}[h]
\centering
\begin{tabular}{ | p{0.5\linewidth} | c | c |} \hline
Criterion & Accuracy & $n$ \\ \hline\hline
SOTA & 93.0 $\pm$ 5.50  & 100 \\
Can reason & 96.0 $\pm$ 4.22 & 100 \\
Cannot reason & 100.0$\pm$ 0.00 & 100 \\
Emergent behaviour & 83.0 $\pm$ 8.10  & 100 \\
Superhuman capabilities    & 97.0 $\pm$3.68 & 100 \\ \hline
Limitations section & 93.0 $\pm$5.50 & 100 \\
Ethics section      & 97.0 $\pm$ 3.68 & 100 \\
Negative results section & 95.0 $\pm$ 4.70  & 100 \\ 
Error breakdown       & 88.0 $\pm$ 7.01 & 100 \\ \hline
Versions  & 82.0 $\pm$ 8.28 & 100 \\
Call Parameters & 86.0 $\pm$ 7.48 & 100 \\
Account for Randomness & 90.0 $\pm$ 6.47 & 100 \\
Open-Sourcing & 74.0 $\pm$ 9.46 & 100 \\ \hline
Statistical tests  & 89.0 $\pm$ 6.75 & 100  \\
Non-English eval.   & 98.0 $\pm$ 3.02 & 100 \\
Dialect eval.    & 100.0 $\pm$ 0.00 & 100 \\ \hline
Human evaluators & 89.83 $\pm$ 8.51 & 59 \\
LLM evaluators & 88.54 $\pm$ 7.01 & 96 \\
Automatic evaluators & 99.51 $\pm$ 1.05 & 204 \\ \hline
Type of text & 99.0 $\pm$ 2.14 & 100 \\ 
LLM-as-subject & 89.0 $\pm$ 6.75 & 100 \\ \hline
Total        & 91.91 $\pm$ 1.22 & \\ \hline
\end{tabular}
\caption{Accuracy for the model for a 95\% interval with sample size ($n$). 
Given that some papers did not contain the criteria we evaluated (e.g., the LLM-as-an-evaluator metric is rare in papers prior to 2023), or were too skewed (dialects stands out on this), we sampled extra for these--hence the overcounting in automatic evaluations. 
Overall, the model performs well as a labeller. Some criteria were better-performing than others (e.g. open-sourcing versus SOTA claims). 
We attribute this to prompting and model capabilities: close inspection of the papers noted that the model tended to overlook content, sometimes verbatim, matching the criteria. 
}\label{tab:cieval}
\end{table}
\end{center}

\section{Extended Results}\label{app:extended}

\subsection{Corpus Composition}\label{app:corpuscomposition}
\begin{figure}[th]
    \centering
    \includegraphics[width = \linewidth]{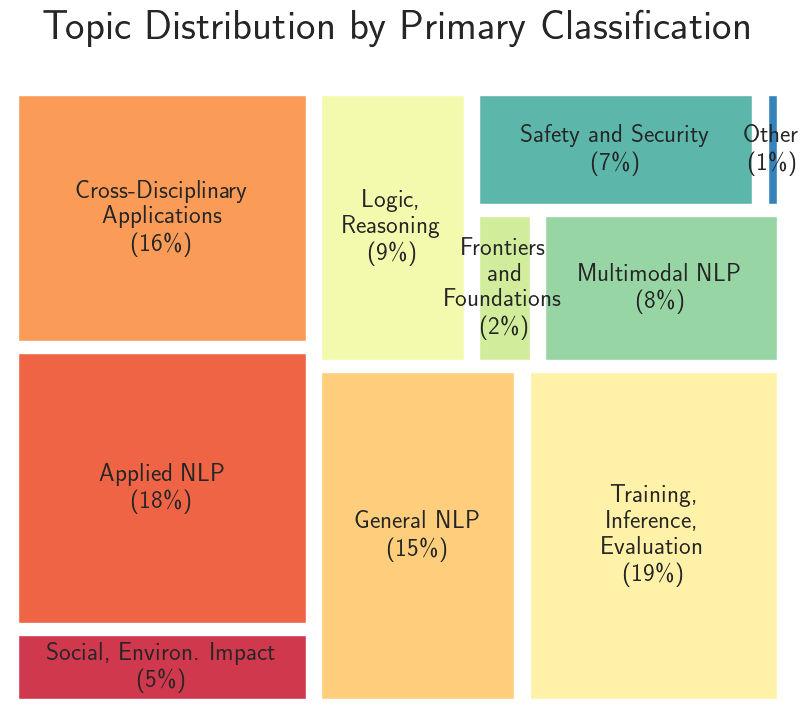}
    \caption{Most common topics in LLM literature, distributed by primary subject. 
    There is a relatively even distribution between cross-disciplinary applications; training, inference, and evaluation methods; applied NLP and general NLP. 
    Within cross-disciplinary applications, these are overwhelmingly in favour of software, and medicine and healthcare.
    }
    \label{fig:topicdistribution}
\end{figure}

\begin{figure}[th]
    \centering
    \includegraphics[width = \linewidth]{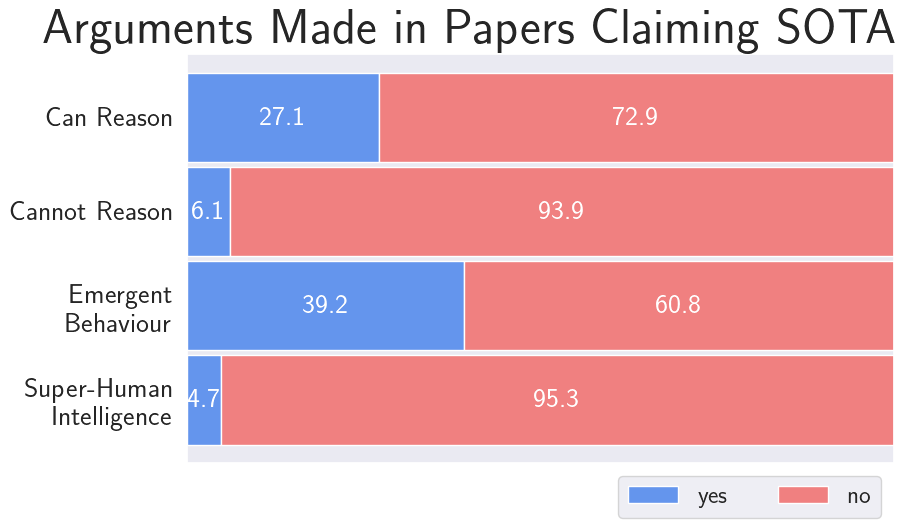}
    \caption{Arguments Made for the papers claiming SOTA results. 
    Some of the arguments made, especially emergent behaviour, usually showed a lower prevalence of structural features considered to be good research. %
    }
    \label{fig:sotaall}
\end{figure}

\begin{figure}[th]
    \centering
    \includegraphics[width = \linewidth]{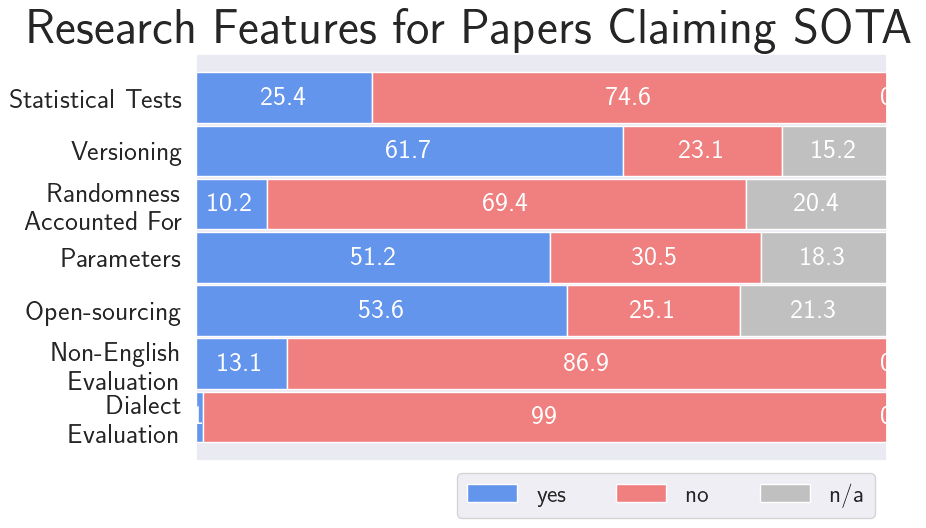}
    \caption{Research quality for relevant papers claiming SOTA. 
    Certain research features, such as non-English and dialect evaluations are very low, but various research protocols, such as open-sourcing, versioning, and paper declarations are more frequently seen. 
    }
    \label{fig:sotarq}
\end{figure}

In this section we present the full results of corpus composition results: percentage-wise (\figstworef{sotaall}{sotarq}) and change over time (\figref{sotarq}). 
We also show our results of the topic analysis work. 
Most topics had a relatively even distribution (\figref{topicdistribution}), but there was a clear focus on applications and improvements, as opposed to safety or social and environmental impact. 
This is not indicative of a problem, however, as we noted in \secref{yearly} that these topics are on the rise. 
It is likely that this disparity in volume is propped up by cross-disciplinary applications. 

\begin{figure}[th]
    \centering
    \includegraphics[width = \linewidth]{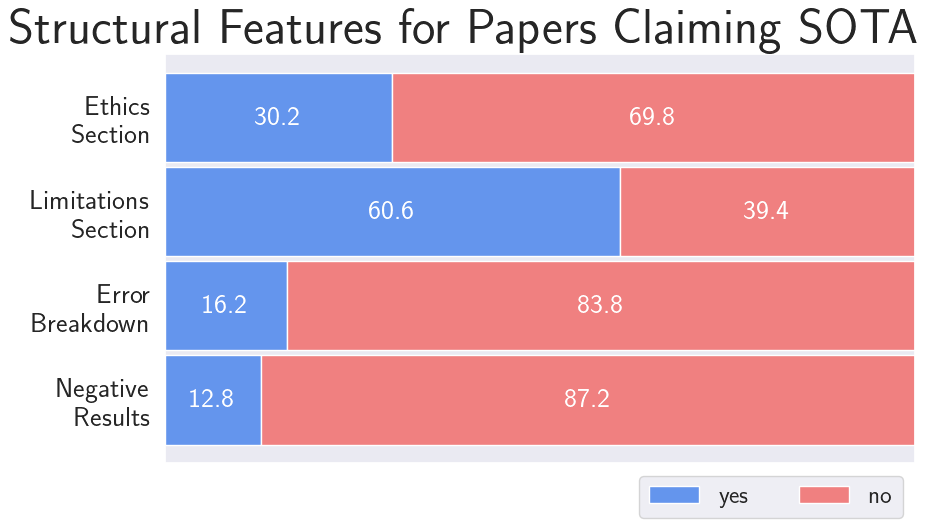}
    \caption{Structural Features for the papers claiming SOTA results. 
    Overall we found a low prevalence of papers containing ethics disclaimers, error breakdowns, and negative results. 
    }
    \label{fig:sotaall2}
\end{figure}

\subsection{Research Features and Types of Evaluator}\label{app:composition}

\begin{figure*}
    \centering
    \includegraphics[width=\linewidth]{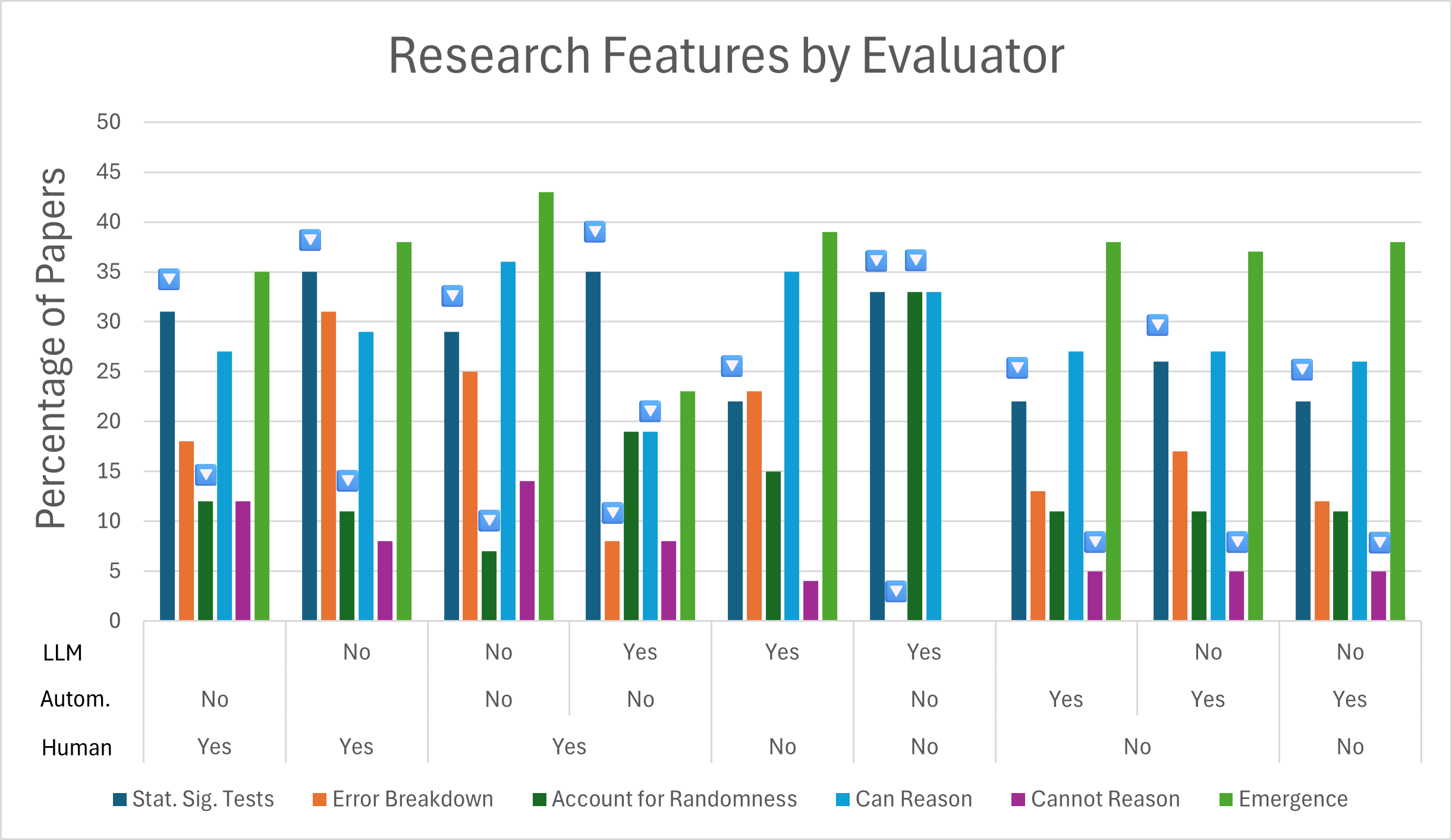}
    \caption{
Percentage breakdown of Research Features (experimental protocols) for various subsets of evaluators, for papers claiming SOTA. Marked with an arrow pointing down are metrics lower than the general corpus by $<$ 2\%. When compared to the general corpus, SOTA papers used fewer measures of statistical significance, and relied more in LLM evaluators. That said, papers \textit{only} LLMs as evaluators were exceedingly rare, and mostly related to papers claiming emergent behaviour. 
}\label{fig:featuresperevaluator}
\end{figure*}

Results broken down by Research Feature and type of evaluator are in \figref{featuresperevaluator}. 
Papers claiming SOTA and relying solely in LLM evaluators were exceedingly rare. 
A considerable amount of papers relied on either only automatic evaluations (55\%) or human evaluations without LLMs (23\%). Relying on one type of evaluator was rare for humans (2\%) and LLMs (statistically insignificant). 
Of note, a large portion of the papers relying on LLM evaluators without humans (8\%) presented error breakdown analyses (23\%; compare with automatic evaluation subsets at 13-17\%). Claims of reasoning were often done with LLM evaluators and not human evaluators (35\%); contrasting with claims that they cannot reason predominantly done with human evaluation alone (14\%). 
When comparing with the entire corpus, SOTA papers use fewer measures of statistical significance across the board. It is also predominant the use of automated metrics (+10\% on average) and fewer Human-only and Human and Automatic only metrics (from 6\% to 2\% and 8\% to 4\%). 
On the other hand, LLM evaluators remained steady, showing that most of the papers claiming SOTA used this metric in some form. 

\subsection{Yearly Topic Changes}\label{app:yearlytopics}

\begin{figure*}
    \centering
    \includegraphics[width=\linewidth]{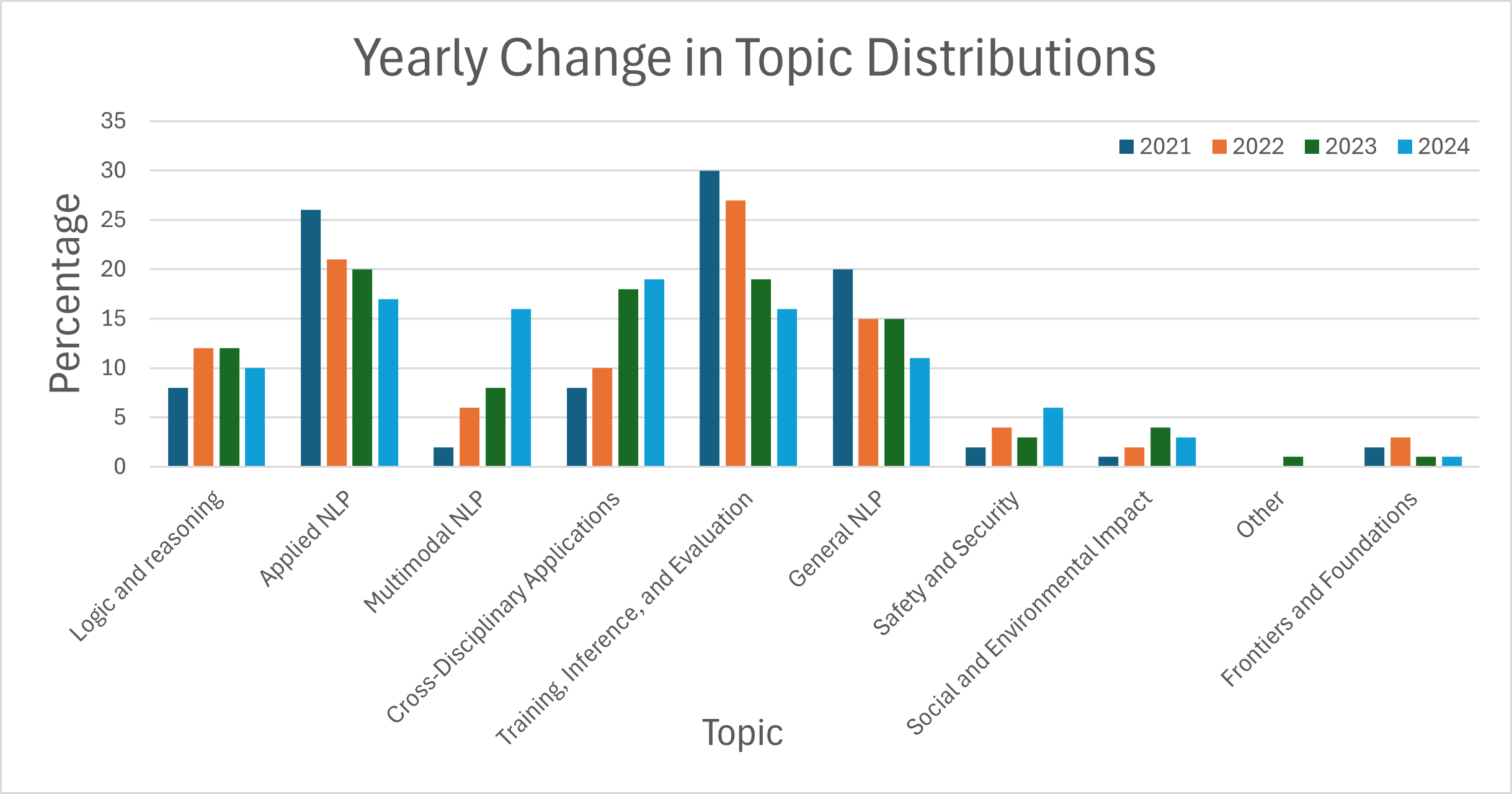}
    \caption{Yearly change in topic distributions for papers focusing on LLMs. 
We noticed a yearly increase in the volume of papers that involved multimodality, security and safety, and cross-disciplinary applications, which coincides with a more widespread adoption of this technology. }
    \label{fig:yearlytopic}
\end{figure*}

The yearly topic changes can be found in \figref{yearlytopic}. While there is a decrease on general NLP--likely attributed to the shift from LLMs as research subjects to research tools--there are considerable increases in other areas, such as cross-disciplinary and multimodal applications. Safety and security appeared to also be modestly rising. 

\subsection{Criteria versus Citations Analysis}\label{app:kolmogorovsmirnov}

To determine whether the presence of a criterion impacted the number of citations, we used a Kolmogorov-Smirnov test. This test is suitable for the task because it is non-parametric, and hence more robust to priors, at the expense of needing larger data sizes. 

Concretely, the null hypothesis $H_{0}$ in this test is that both samples come from the same underlying distribution. 
Accepting (rather, being unable to reject) $H_{0}$ means that the distributions are statistically indistinguishable, and we are \emph{unable to conclude} that likelihood of citation is impacted by the presence of the criterion. 
Rejecting $H_{0}$ implies that the distributions are distinct, and we may conclude that the criterion \textit{does} impact the number of citations received. 

Given a calibrated $p$-value and samples of lengths $m$ and $n$, the Kolmogorov-Smirnov test condition is given by
\begin{equation}
    D_{m, n} = \sqrt{-\ln{(p/2)} \cdot (\frac{n + m}{2mn})}. 
\end{equation}
If the test statistic (percentage citation difference) $D$ is $D > D_{m, n}$, we reject $H_{0}$. We calibrated our $p$ value to $p< 0.05$. An interpretable version of this calibration is that we expect to be wrong about our conclusions 5\% of the time. 
We capture our results in \tabref{kstable}. Overall, we deemed that 7 out of the 18 criteria did not impact the number of citation number. 
Of note these were the use of LLMs as evaluators, open sourced artifacts, and the presence of ethics and limitations sections. 

\begin{center}
\begin{table}[ht]
\centering
\begin{tabular}{ | c| c |} \hline
Criterion & $H_{0}$ \\ \hline\hline
Statistical Tests &  Accept \\
\cellcolor{blue!15}Version Declaration & \cellcolor{blue!15} Reject \\
Parameter Declaration &    Accept \\
Account for Randomness &   Accept \\
Non-English Evaluation &   Accept \\
Dialect Evaluation &   Accept \\
\cellcolor{blue!15}Open Sourcing &   \cellcolor{blue!15} Reject \\
\cellcolor{blue!15}LLM Evaluators &   \cellcolor{blue!15} Reject \\
Human Evaluators &    Accept \\
\cellcolor{blue!15}Automatic Evaluators &  \cellcolor{blue!15} Reject \\\hline
\cellcolor{blue!15}Limitations Sections & \cellcolor{blue!15}  Reject \\
\cellcolor{blue!15}Ethics Sections &  \cellcolor{blue!15} Reject \\
Negative Results &  Accept \\
Error Breakdowns &  Accept \\\hline
Emergence Claims &  Accept \\
\cellcolor{blue!15}Can Reason Claims & \cellcolor{blue!15} Reject \\
Cannot Reason Claims &  Accept \\
Super-Human Capability Claims &   Accept \\\hline
\end{tabular}
\caption{Impact of the presence of a given criterion on its citation number. 
To determine this we split the distribution into samples containing and not containing the criterion and ran a Kolmogorov-Smirnov test. 
Rejecting $H_{0}$ (in blue) implies that the presence of the criterion does \textit{not} impact the citation number. }\label{tab:kstable}
\end{table}
\end{center}

\subsection{Yearly Changes}\label{app:yearlychanges}

\begin{figure*}
    \centering
    \includegraphics[width=0.9\linewidth]{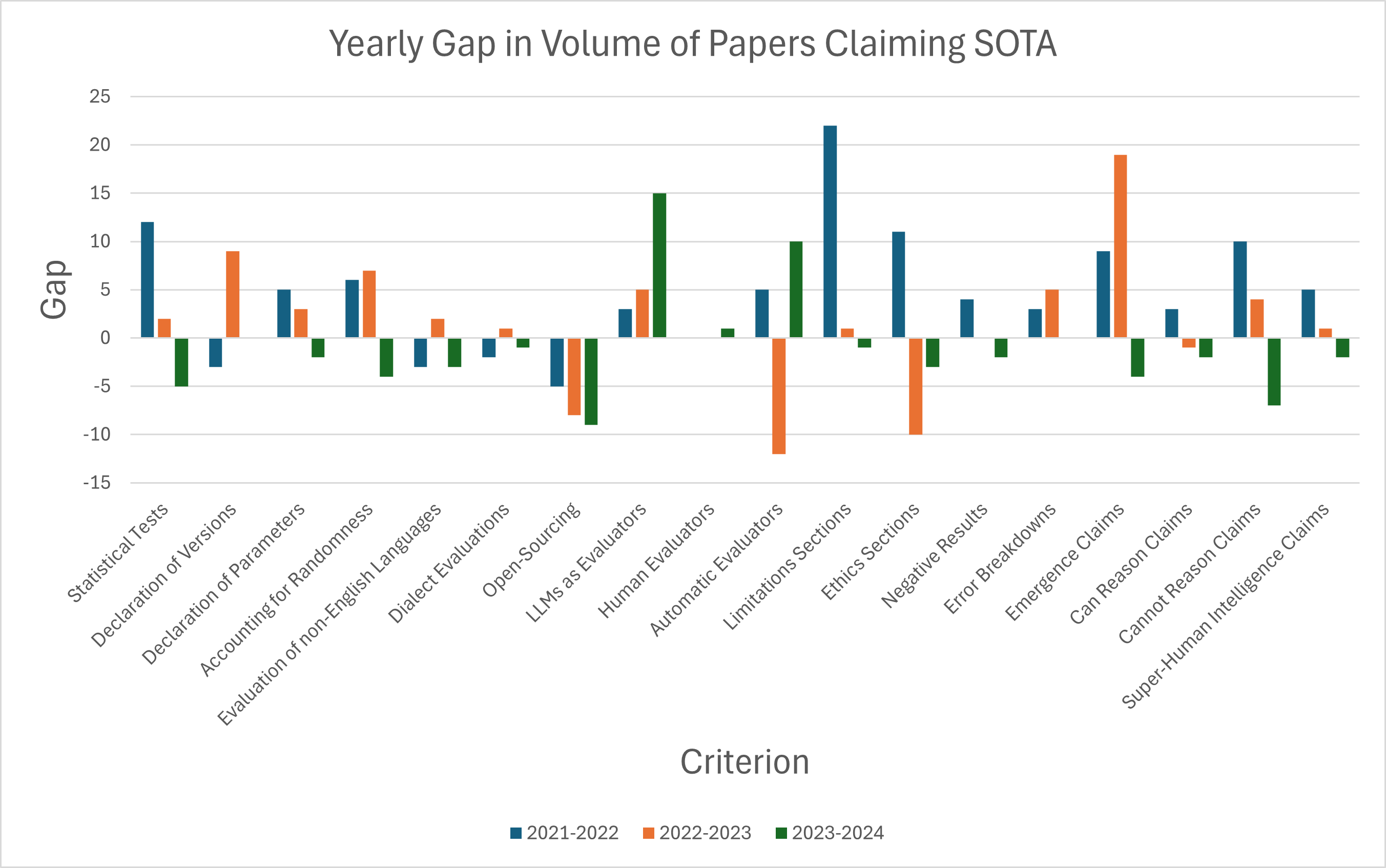}
    \caption{Yearly percentual change (gap) in volume of papers claiming SOTA and presenting the given criterion as an absolute percent. 
This quantity is more interpretable in this scenario: for example, the volume of papers claiming reasoning capabilities in 2023 was 103 out of 294 papers, or 35\%. In 2024 this number was 165 out of 535, or 31\%. 
The gap is then -4\%. 
Unlike in \figref{yearlypercentagecitations}, the percentages for 2024 are not dependent on their recency: 2024 amounts for 46\% of the papers evaluated. 
We observed decreases in all trends, except versioning and all evaluators.}
    \label{fig:yearlypercentagevolume}
\end{figure*}

\begin{figure*}
    \centering
    \includegraphics[width=0.9\linewidth]{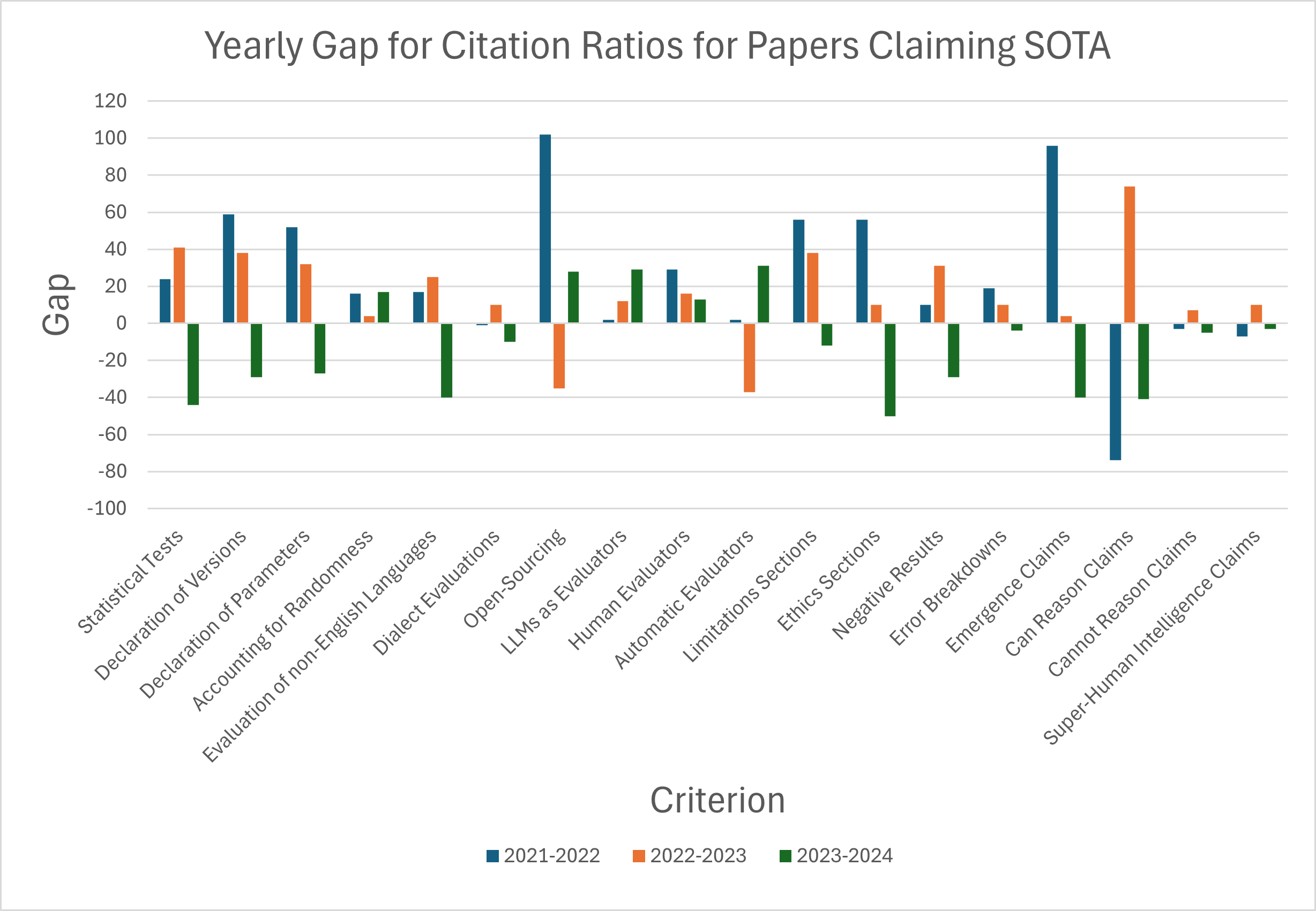}
    \caption{Yearly absolute percentage changes for citation ratios (gap) for all our criteria across the years, for papers claiming SOTA. 
Papers with human evaluators had 7,793 citations in 2022, versus 26,653 without. In 2023, this number was 12,592 (with) and 28,243 (without). 
The gap is then -54\% in 2022 and -38\% in 2023, and the change in the gap is +16\%. 
Note that 2024 accounts for most of the papers in our corpus (46\%) but has the lowest number of citations (7\%) due to recency bias: hence, most of the gap drops are expected.}
    \label{fig:yearlypercentagecitations}
\end{figure*}

 We show yearly trends in the gap for the volume of papers claiming SOTA (\figref{yearlypercentagevolume}) and for the citation ratios (\figref{yearlypercentagecitations}) per criteria. 
The core findings of this section are that in terms of volume (recall that 2024 comprises almost half of our corpus), there was a decrease in papers having statistical tests, claims of emergence and reasoning, and open sourcing. There were, however, increases in the use of LLMs as evaluators. Some other criteria remained steady, such as dialect evaluations. This metric is not susceptible to recency bias. 
On the other hand, there \emph{is} a recency bias in the citation ratios, since, by definition, this is a time-dependent metric, and so drops in 2024 are expected. 

\section{Future Trends in Citation  Volume}\label{app:googlescholar}

Our work parted from the assumption that most LLM-related works would cite at least one of the GPT-3 or GPT-4 papers. 
This assumption must be examined closely, and in particular be put in contrast with future trends like the rise of--comparatively--open models such as LlaMA. 

For this, we performed a follow-up study where we examined the citation volume a year after the cutoff date for the data used in this paper. 
Overall, most models have not reached the citation volume of the original GPT-3.5 or GPT-4 works. However, there exist exceptions to this rule: the first LlaMA paper, released at the same time as the GPT-4 technical report, has over 5,000 more citations than GPT-4's; but about a third of GPT-3's. 
In comparison, most papers have up to a third of GPT-4's (Gemini, with 4,200; \citealt{geminiteam2025geminifamilyhighlycapable}). 
The LlaMA paper is impactful, however: 
remark that the GPT-3 paper is three years older. Three years, as per our results, symbolises a significant amount of relative time in this field. 

Even though it is very likely some articles could progressively stop citing both GPT papers, we expect this trend to be farther in the future. 
Likewise, it is very possible that some, if not most, of the citations in LlaMA papers include also references to at least one of the GPT works.
Hence, our core assumption holds--to a degree, given the omission of the LlaMA paper in our work. 

\begin{figure*}
    \centering
    \includegraphics[width=\linewidth]{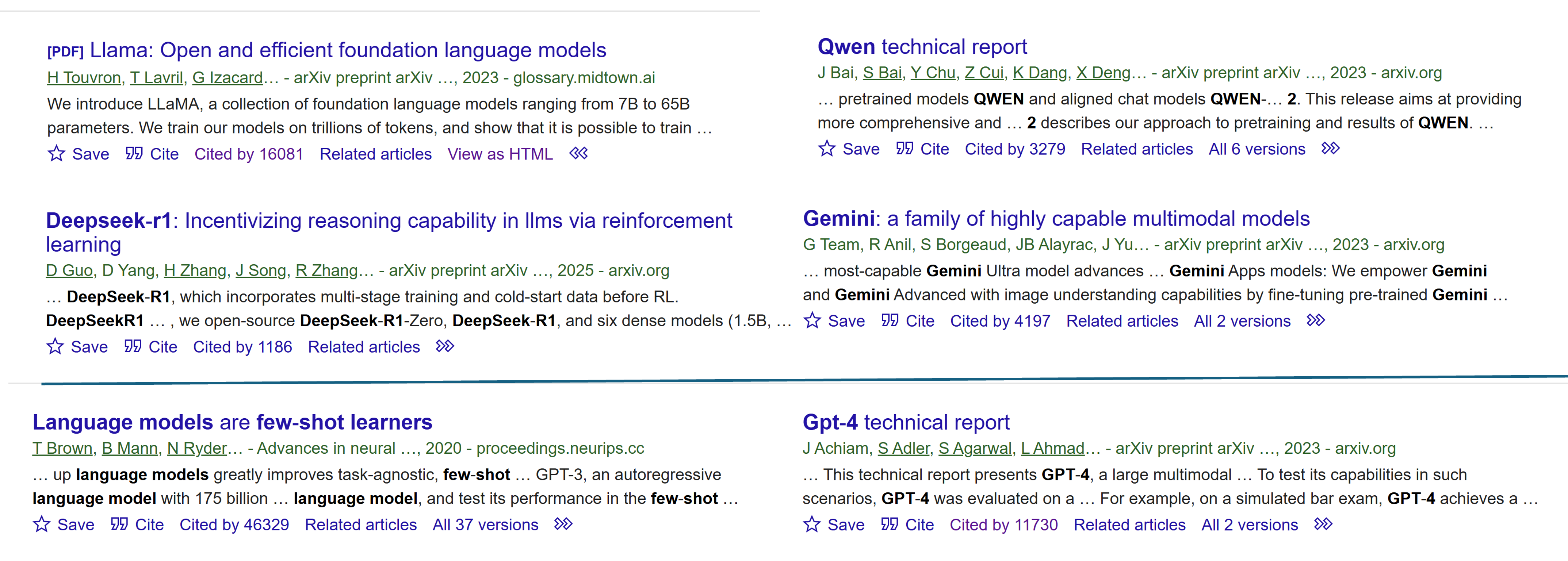}
    \caption{Citation volume from (clockwise from the top left) the LlaMA, Qwen \cite{bai2023qwentechnicalreport}, Gemini, GPT-4, GPT-3, and Deepseek \cite{deepseek} models, as of 27 May 2025. 
    While most models do not reach the volumes of the papers studied in this work, there is a definite increasing trend in some LLM-related works. In particular, the Llama paper has 5,000 more citations than GPT-4's technical report, in spite of both papers being released the same year.}
    \label{fig:futuretrends}
\end{figure*}

Unfortunately, at the time of writing this, most public APIs, including the Internet Archive's Wayback Machine and Publish or Perish, were no longer able to query Google Scholar. 
This in turn hindered our ability to perform a comprehensive follow-up. 
Instead, we attach screenshots for the citation trends in \figref{futuretrends}. 

\end{document}